# Assessment of Conditional Diffusion Model for Synthetic Histopathology Image Generation

Seyed Kahaki[*], Shijie Li, Weijie Chen, Nicholas Petrick
Division of Imaging, Diagnostics, and Software Reliability, U.S. Food and Drug Administration (FDA)

## ABSTRACT

Synthetic histopathology image generation has emerged as an approach that may address data scarcity in computational pathology, yet current evaluation methodologies may not fully assess synthetic data quality for medical applications. This work investigates and addresses limitations in existing evaluation metrics, investigating an approach for assessing synthetic histopathology image quality through domain-specific metrics and downstream task validation. We show that conventional synthetic data evaluation metrics such as Fréchet Inception Distance (FID) and Inception Score (IS) may have limitations when applied to histopathology images due to their reliance on ImageNet-pretrained feature extractors. To address these limitations, we propose for consideration modified FID and IS approaches utilizing foundation models pretrained on digital pathology datasets (UNI, Virchow, Virchow2), supplemented by precision-recall based metrics as part of an additional quality assessment. Using conditional denoising diffusion models trained on four benchmark datasets (MoNuSeg, TNBC, 2018 Data Science Bowl, PanNuke), with a two-step training approach, we generated synthetic datasets with systematically varied quality characteristics. We also measured the correlation between the synthetic data quality metrics with downstream nuclei segmentation performance using common metrics including the aggregated Jaccard index (AJI+) and the Dice coefficient. The study results suggest that pathology-specific metrics may provide improved discriminative power. Specifically, the modified Inception Score indicates higher correlation with downstream task performance ($r = 0.6096$ with AJI+, $p = 0.0122$), compared to the original IS ($r = 0.0708$, $p = 0.7944$). Our observations indicate that increasing the variety of generated training data has a higher positive correlation with segmentation model performance than improving the visual fidelity of individual generated images. This research proposed a domain-specific evaluation approach and further suggests a potential relationship between the modified synthetic data performance metrics and downstream task performance.

**Keywords:** Synthetic data quality, histopathology, image-to-image translation, denoising diffusion model

* Corresponding author: seyed.kahaki@fda.hhs.gov

## 1. INTRODUCTION

The development of AI models in digital pathology faces a notable challenge: the scarcity of annotated whole-slide images (WSIs) and the prohibitive cost of expert pathologist annotations [1-5]. While synthetic data generation has emerged as a promising solution to augment limited training datasets, a gap exists in current methods for evaluating the quality and utility of synthetically generated histopathology images. Conventional evaluation metrics may not capture the domain-specific characteristics relevant to medical image assessment. Fréchet Inception Distance (FID) [6] and Inception Score (IS) [7], current synthetic data quality metrics, rely on InceptionV3 feature extractors pretrained on ImageNet [8] which is a mismatch with medical imagery. This results in compressed score ranges when applied to histopathology image data which may limit discriminative power for histopathology-specific quality differences. To address these fundamental evaluation challenges, we proposed modified data quality metrics specific to histopathology. We also present the correlation between these synthetic data quality metrics with downstream nuclei segmentation performance using common metrics including AJI+ and Dice coefficient via the HoVerNet model [9]. Generative models have evolved from Variational Autoencoders (VAEs) [10] to Generative Adversarial Networks (GANs) [11] and Denoising Diffusion Probabilistic Models (DDPMs) Ho, Jain and Abbeel [12]; Sohl-Dickstein, Weiss, Maheswaranathan and Ganguli [13]. While GANs have shown notable success in certain digital pathology applications, [14, 15], they suffer from training instability and mode collapse, particularly problematic for capturing rare morphological features [16]. DDPMs address these GAN limitations through stable training and superior mode coverage [12, 13], making them potentially more applicable for histopathology data generation [2]. Using conditional denoising diffusion models [12] based on DDPMs trained on four benchmark histopathology datasets (MoNuSeg [17, 18], TNBC [19], 2018 Data Science Bowl [20], PanNuke [21]), we generated synthetic datasets with systematically varied quality characteristics with a two-step approach.

This paper makes the following contributions:

- the introduction of a two-step training approach for denoising diffusion models,
- the development of domain-specific modified FID and IS metrics using pathology-pretrained foundation models (UNI [22], Virchow [23], Virchow2 [24]) supplemented by precision-recall metrics, and
- an empirical validation approach that includes correlation analysis between the modified synthetic data evaluation metrics and downstream task-specific nuclei segmentation performance including dice Similarity Coefficient (DICE) and *Improved Aggregated Jaccard Index (AJI+)*.

## 2. MATERIALS AND METHODS

The first phase of our study focused on generating synthetic histopathology images using conditional Denoising Diffusion Probabilistic Models (DDPMs) [12] in a two-step approach. To achieve this, a separate DDPM was trained using the training and tuning data from each of the four distinct histopathology datasets (MoNuSeg, TNBC, 2018 Data Science Bowl, and PanNuke). The conditioning mechanism, utilizing instance nuclei segmentation masks, was implemented to guide the image generation process, to guide generation of histopathological features consistent with the conditioning masks. Following the training of each model, synthetic histopathology images were generated by providing nuclei segmentation masks from the corresponding test partition from each of the four datasets as conditional inputs.

The second phase of our study involved developing modified synthetic data quality metrics specific to histopathology. We adopted the 7C assessment approach proposed by Zamzmi et al. [25] to evaluate the synthetic images. In this work we focus on two key dimensions: congruence (fidelity which measures how realistic the synthetic image distribution is and similarity which measures how closely synthetic images match real images at the distribution level between synthetic and real image distributions) and coverage (the variety of the generated data) [25]. While congruence is commonly assessed in generative modeling, coverage is often underemphasized [25, 26]. However, coverage is particularly important when synthetic data is intended to augment training sets for downstream tasks such as nuclei segmentation because high variety in the generated images may potentially enhance the robustness of the downstream tasks. This included modifying the traditional FID and IS implementations by utilizing the UNI, Virchow, and Virchow2 foundation models pretrained specifically on the digital pathology datasets supplemented by precision-recall based metrics [27, 28].

Finally, the subsequent performance evaluation, detailed in the results section, was systematically conducted by comparing the generated synthetic images against the ground truth images from the respective test sets. This was accomplished by employing a suite of quantitative metrics and qualitative pathologist assessment.

### Synthetic Data

#### *Generative Model Structure and Training*

The model architecture is a U-Net architecture, conditioned on instance segmentation masks [19]. These masks are structured with three channels to provide detailed spatial and instance-level guidance for the generation process. The first channel is a binary foreground-background mask, distinguishing nuclear regions from the background. The second channel represents a horizontal map for each nucleus, where pixel values range from -1 (leftmost extent of a nucleus) to 1 (rightmost extent), with intermediate pixels assigned values linearly interpolated according to their horizontal position within the nucleus. Similarly, the third channel provides a vertical map, with pixel values ranging from -1 (upmost extent) to 1 (bottommost extent) of a nucleus, and intermediate pixels interpolated based on their vertical position. The conditioning information from these detailed segmentation masks is integrated into the U-Net using a spatially adaptive normalization (SPADE) module [29]. The SPADE module utilizes the input masks to modulate the activations within the U-Net's normalization layers, thereby spatially tailoring the generated features to align with the provided nuclear structures (see Figure 1). The multi-channel nature of these masks, particularly the horizontal and vertical gradient maps, allows the SPADE module to leverage rich instance-specific spatial information. We first train the U-Net following the set-up of [2] with a learning rate of 0.0001. After convergence, we train the network without a classifier and set the learning rate to 0.00002. The segmentation mask is injected into the U-Net using a spatially adaptive normalization (SPADE) module [29]. To systematically generate synthetic datasets with varied quality characteristics, we employed a two-step training approach for the conditional denoising diffusion models:

Coarse Model Training: We first trained the U-Net following the setup of [2] with a learning rate of 0.0001 until convergence. This initial training phase produces a "coarse" model that generates synthetic images with adequate structural coherence but may exhibit color bias and reduced texture realism.

Fine-Tuned Model Training: After the coarse model converged, we continued training without a conditional training component and reduced the learning rate to 0.00002. This fine-tuning phase produces a "fine-tuned" model that generates images with improved color fidelity and enhanced textural details that more closely match real histopathology samples.

This two-step approach allows us to generate synthetic datasets at different quality levels, enabling systematic evaluation of how synthetic data quality characteristics correlate with downstream task performance. Both coarse and fine-tuned models were trained on each of the four benchmark datasets (MoNuSeg, TNBC, 2018 Data Science Bowl, PanNuke).

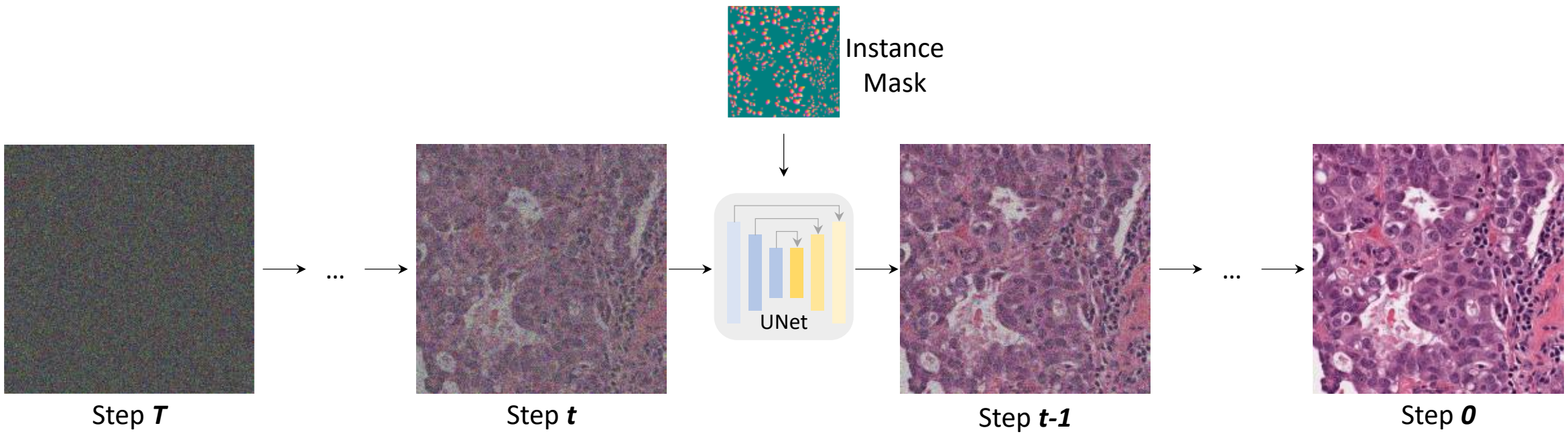

Figure 1. Illustration of the Conditional Denoising Diffusion Model for generating synthetic histopathology images. The generation process begins by initializing the image with random noise, where each pixel in each color channel follows a Gaussian distribution (e.g., $x_T$). This initial noise image is then iteratively refined by the trained U-Net over a predefined number of denoising timesteps (e.g., T steps). At each timestep t (from T down to 1), the U-Net conditioned on the instance nuclei segmentation mask, predicts the noise component present in the image $x_t$ (or, alternatively, predicts the less noisy image $x_{t-1}$). This iterative denoising process progressively transforms the initial random noise into a coherent synthetic histopathology image that corresponds to the conditioning segmentation mask.

### *Dataset Description*

The datasets utilized in this study were selected to provide a range of histopathological features, imaging conditions, and cell types, enabling a comprehensive evaluation of the synthetic image generation model and evaluation metrics. The MoNuSeg dataset [17, 18], Hematoxylin and eosin (H&E) stained tissue images at 40x magnification from The Cancer Genome Atlas Program (TCGA) archive, includes annotated images from multiple patients with various organ tumors diagnosed from different hospitals. We use 30 MoNuSeg images (22,000 nuclei) for training and tuning and 12 images for testing. Images are of size 1000x1000.

The Triple-Negative Breast Cancer (TNBC) dataset [19] contains annotations for a broad spectrum of cell types, such as normal epithelial and myoepithelial cells, invasive carcinomatous cells, and various immune cells. We use 40 TNBC images for training and tuning and 10 images for testing, totaling 4022 annotated cells. Images are size 512x512.

The 2018 Data Science Bowl dataset [20] offers a variety of segmented nuclei images under differing conditions, challenging the model's ability to generalize across cell types, magnifications, and imaging modalities. It contains H&E, fluorescent, and bright field images. In this study, we only use the H&E-stained images. We use 670 images for training and tuning, and 65 images for testing. This dataset includes images of varying sizes from 256x256 to larger.

The PanNuke Dataset [21] is an open-access, pan-cancer histology dataset designed for nuclei instance segmentation and classification. It contains a wide variety of images from different cancer types, providing a disparate set of histopathological samples. The dataset includes images of tissue samples stained with H&E, and features both tumor and non-tumor regions from multiple organs. The PanNuke dataset is particularly challenging due to its variability in tissue types, image resolutions, and staining protocols. It includes 8,000 images across 10 different cancer types, with annotations for nuclei segmentation and classification tasks. In this study, we use a subset of 6,000 images for training and validation, and 1,000 images for testing, ensuring robust evaluation of model performance across different cancer types and tissue structures. Each image is size 256x256, typically 40x magnification (with some source images originally scanned at 20x and interpolated to a pixel size equivalent to 40X for consistency).

Experiments utilized the data splits suggested in the original publications of these datasets: MoNuSeg (37/14/14 Training/Tuning/Testing images), TNBC (22/18/10), 2018 DSB (108/12/21), and PanNuke (827/749/775). Training data was kept separate; each synthetic generator and segmentation model was trained and tuned using only the real and synthetic data from its specific dataset.

### *Synthetic Data Evaluation Metrics*

We evaluate the synthetic data along the congruence and coverage dimensions of 7C approach [30]. For the assessment of congruence, particularly when evaluating image sets from generative models, we utilize Fréchet Inception Distance (FID), Structural Similarity Index Measure (SSIM), Peak Signal-to-Noise Ratio (PSNR), and Precision. For evaluating coverage, we employ the Inception Score and Recall metric. Regarding the downstream segmentation task, performance is evaluated using established metrics. The Dice Score is employed to measure semantic segmentation accuracy, while the Aggregated Jaccard Index (AJI+) is used to assess instance segmentation quality.

**Congruence** The performance evaluation was conducted by comparing the DDPM synthetic images with the real testing data. We used the FID [6] to assess the visual quality and realism of the synthetic images. The FID is calculated as Eq. 1:

$$FID = |\mu_r - \mu_g| + Tr\left(\Sigma_r + \Sigma_g - 2\left(\Sigma_r\Sigma_g\right)^{1/2}\right) \tag{1}$$

where $\mu_r$ and $\mu_g$ are the mean feature vectors of real and generated samples, respectively. The feature vectors are extracted based on a Inception-v3 convolutional neural network pre-trained on ImageNet dataset [8]. $\Sigma_r$ and $\Sigma_g$ are the covariance matrices of the features vectors of real and generated samples respectively. $|.|$ denotes the Euclidean distance (L2 norm). $Tr$ denotes the trace of a matrix. Besides FID, we also calculate Precision (P). Formally, for a generated sample distribution $G$ and a real data distribution $R$ with manifolds $M_G$ and $M_R$ respectively, precision can be expressed as:

$$P = \frac{|\{x \in M_G: \exists y \in M_R, d(x, y) < \varepsilon\}|}{|M_G|} \tag{2}$$

We also evaluated Structural Similarity Index Measure (SSIM) [31] and Peak Signal-to-Noise Ratio (PSNR) to assess pixel-level similarity and signal quality between synthetic and real images. SSIM values range from -1 to 1 (with 1 indicating identical images), while PSNR is measured in decibels (dB), with higher values indicating better quality.

**Coverage** The Inception Score (IS) [7] was used to evaluate the variety of the synthetic dataset, ensuring it captures a wide range of histopathological variations. The IS is calculated as:

$$\text{IS} = \exp\left(E_{\mathbf{x}}\left[\text{KL}\left(p(y|\mathbf{x})|p(y)\right)\right]\right) \tag{3}$$

where $p(y|\boldsymbol{x})$ is the conditional probability distribution of the label $y$ (a feature vector output from the network model) given the generated image $\boldsymbol{x}$. $p(y)$ is the marginal probability distribution of the label (i.e., the expectation of p(y|x) over all the testing images). $KL(\dot{}|\dot{})$ denotes the Kullback-Leibler (KL) divergence. $E[.]$ represents the expectation over the generated images. In this case, $p(y|\boldsymbol{x})$ is a predicted probability distribution given a generated image $\boldsymbol{x}$, while $p(y)$ is the average predicted probability distribution over all the generated images. Recall (R) is mathematically defined as:

$$\text{R} = \frac{|\{\text{y} \in \text{M}_\text{R}: \exists \text{x} \in \text{M}_\text{G}, \text{d}(\text{x}, \text{y}) < \varepsilon\}|}{|\text{M}_\text{R}|} \tag{4}$$

where d(x, y) is Euclidean distance metric between a generated sample x and a real sample y, with ε is a distance threshold. $M_G$ is the manifold of generated (synthetic) samples, the structured space where all the synthetic images lie, and the $M_R$ is the manifold of real data samples, the structured space where all the real histopathology images lie. Precision and Recall values range from 0 to 1, with higher values indicating better performance. Precision values above 0.5 suggest that the majority of generated samples fall within the real data manifold, while Recall values above 0.5 indicate good coverage of the real data distribution [27, 28].

### *Domain-Specific Evaluation Metrics*

To address the limitations of InceptionV3, which is pretrained on natural images, we developed modified versions of the FID, IS, Precision, and Recall metrics using feature extractors from pathology-specific foundation models: UNI, Virchow, and Virchow2. We calculated both the original InceptionV3-based metrics and the modified pathology-specific metrics for all datasets to enable direct comparison. For each feature extractor, we preprocessed images and extracted features following a consistent pipeline. All images were divided into non-overlapping patches to align with the original training methodologies of each model: 224×224 pixel patches for InceptionV3 [32] and 256×256 pixel patches for UNI, Virchow, and Virchow2 [24, 32]. For all approaches, features were extracted from individual patches and then aggregated by averaging the feature vectors across all patches within each image to obtain a single feature vector representation per image. For InceptionV3, features were extracted using the standard approach without patch-level aggregation. The metrics were then calculated using the same sets of real and synthetic histopathology images across all feature extractors to ensure fair comparison.

### *Feature vectors extracted by different feature extractors*

To interpret the behavior of the different feature extractors, we applied the three dimensionality reduction techniques t-SNE, UMAP [33], and PCA [34] to the high-dimensional feature vectors extracted from both the real and synthetic images. We also performed a nearest-neighbor analysis using cosine similarity in the feature space to identify, for each image patch, the most similar image according to each extractor. Finally, a board-certified pathologist performed a qualitative visual review of the synthetic images to assess the presence of histopathological features of interest (carcinoma, stroma, and inflammatory cells).

### Nuclei Segmentation Evaluation

To assess the practical utility of the generated synthetic histopathology images, particularly for augmenting training datasets for common computational pathology tasks, we evaluated their impact on a downstream nuclei segmentation model, specifically HoVerNet [9]. HoVerNet is a deep learning architecture specifically designed for simultaneous nuclear segmentation and classification in histopathology images, utilizing horizontal and vertical distance maps to separate touching nuclei and predict nuclear types. Nuclei segmentation is a foundational step in many digital pathology workflows, as the accurate identification and delineation of nuclei enable quantitative analysis of cellular morphology, spatial arrangements, and cell-type classification. These analyses are important for tasks ranging from cancer grading and prognostication to identifying biomarkers and understanding disease mechanisms. Therefore, the ability of synthetic data to train robust and accurate nuclei segmentation models serves as an indicator of data quality and the potential clinical relevance. For nuclei segmentation evaluation, we used the Dice Similarity Coefficient (DICE) which is a fundamental metric for evaluating the spatial overlap between predicted segmentations and ground truth masks, and AJI+ [35], a modified aggregated Jaccard index metric.

### Correlation with Downstream Task Performance:

To then investigate the relationship between the generative model metrics and the nuclear segmentation metrics, we conducted a correlation analysis between the dataset quality metrics (standard and modified versions of IS, FID, precision, and recall metrics) and the downstream task performance metrics (AJI+ and DICE) on the real and corresponding synthetic datasets.

## 3. RESULTS AND DISCUSSIONS

In this section, we present the synthetic datasets generated by various models (see section 2) under different training configurations (coarse and fine-tuned). Figure 2 illustrates the outputs of the coarsely trained and fine-tuned generative models. Each row presents a different example, consisting of four columns: (1) the nuclei segmentation mask used as the conditioning signal, (2) the corresponding real histopathology image, (3) the image generated by the coarsely trained model, and (4) the image generated by the fine-tuned model. The figure illustrates differences in image quality between coarse and fine-tuned model outputs. While both models produce structurally coherent outputs that preserve the general morphology of nuclei, the coarsely trained model exhibits consistent color bias and reduced realism. This color bias is removed in the images generated by the fine-tuned model, which more closely resembles the color distribution and appearance of real tissue samples.

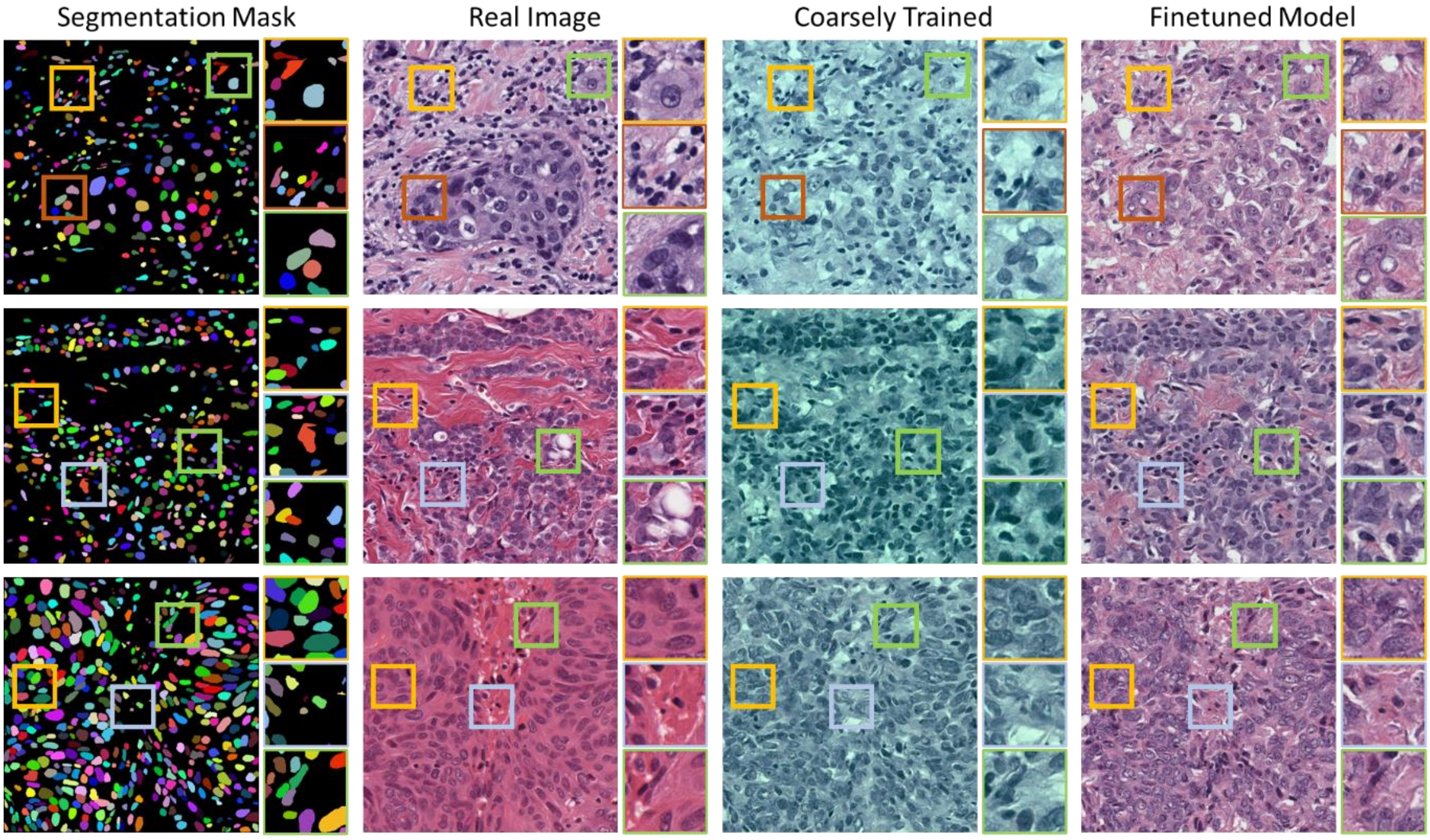

Figure 2. Examples of images generated by the coarse and finetuned models on the MoNuSeg Dataset. The coarsely trained model produces a more blue appearance in images, whereas the finetuned model produces images with a color distribution more akin to the real image.

Table 1 presents the quantitative evaluation of synthetic images using InceptionV3-based metrics. For FID, lower values indicate better fidelity. For IS, values above 1.0 reflect greater predicted label variability, as defined by the metric formulation [7], though the metric shows limited sensitivity with InceptionV3 features. SSIM values range from 0 to 1, SSIM values of 0.98 and PSNR values above 50 dB were observed in this study. Note that Precision and Recall metrics are reported separately in Table 3 for the synthetic datasets. For the fine-tuned model, the average original FID was 83.14, with an average original IS of 1.03, an average SSIM of 0.98, and an average PSNR of 59.83 (Table 1). We observe that the fine-tuned model shows higher fidelity (FID) and higher similarity (SSIM, PSNR) with the real data across all four datasets. However, the variety (IS) does not increase within the fine-tuned model even though visually there are substantive color differences between the coarse and fine-tuned data in the model outputs. It has been reported that GANs may generate high visual fidelity images (observed by pathologist) but with little variety consistent with similar issues reported in the literature [36]. When a segmentation model is trained on low-variety synthetic data, its performance on a disparate real test set may be reduced. In addition to the quantitative evaluation in Table 1, a pathologist review indicates that the generated synthetic data are realistic and contain structures similar to real histopathology data. Specifically, they noted the presence of carcinoma, stroma, lymphocyte on top of carcinoma, inflammatory/lymphocyte in stroma, indicating good realism in the synthetic images.

Table 1. Evaluation Results of the Generated Synthetic Images Across Datasets. The FID metric in this table is the InceptionV3-based. Lower FID values indicate higher fidelity (better match to real data distribution). Higher IS values indicate greater variety in the generated dataset. SSIM values closer to 1.0 indicate higher structural similarity, and higher PSNR values (in dB) indicate better signal quality. The fine-tuned models show improved fidelity (lower FID) and similarity (higher SSIM and PSNR) compared to coarse models across all datasets. ↓ indicated lower value is better, ↑ indicated higher value is better.

| DATASET | | FID↓ | IS ↑ | SSIM↑ | PSNR↑ |
|---|---|---|---|---|---|
| TNBC | Coarse | 70.36 | 1.00 | 0.9881 | 59.00 |
| | Finetune | **67.14** | 1.00 | **0.9941** | **61.92** |
| DSB 2018 | Coarse | 134.69 | 1.05 | 0.9868 | 58.95 |
| | Finetune | **123.64** | 1.05 | **0.9877** | **59.61** |
| MoNuSeg | Coarse | 41.72 | 1.00 | 0.9839 | 57.21 |
| | Finetune | **25.65** | 1.00 | **0.9855** | **57.66** |
| PanNuke | Coarse | 126.77 | 1.07 | 0.9872 | 58.96 |
| | Finetune | **117.14** | 1.07 | **0.9902** | **60.13** |
| Average (Std) | Coarse | 93.85 (38.79) | 1.03 (0.03) | 0.9865 (0.001) | 58.54 (0.76) |
| | Finetune | **83.14** (39.86) | 1.03 (0.03) | **0.9894** (0.003) | **59.83** (1.51) |

## Limitations of Standard Image Quality Metrics

Our analysis revealed some limitations when using InceptionV3-based FID for evaluating synthetic histopathology images. Table 2 shows how the selected model impacts the FID and IS scores. FID scores using InceptionV3 were substantially lower (ranging from 25.65 to 134.69) compared to pathology specific feature extractors (ranging from 481.84 to 2156.59). This marked difference in the dynamic range of FID scores suggests that the InceptionV3 feature space, optimized for natural images, may capture different visual characteristics than those relevant to histopathology and complex textural patterns pertinent to histopathology. The “correct” FID range is not absolute but relative to the feature extractor and dataset; the concern here is the potential for InceptionV3 to be less discriminative for pathology-specific visual features. Furthermore, while the InceptionV3-based FID did show a directional improvement (decrease in score) from coarse to fine-tuned models (e.g., for TNBC: from 70.36 to 67.14; for DSB 2018: from 134.69 to 123.64), the magnitude of this change was relatively small. The limitations of standard metrics reliant on natural image features are even more pronounced with the Inception Score. To adapt IS for the pathology domain, we utilized the normalized feature vectors extracted by the respective networks directly, rather than relying on classification outputs into natural image categories. Despite this adaptation, IS values derived from InceptionV3-based features remained virtually unchanged between real and synthetic images, with scores consistently clustering tightly around 1.00 to 1.07 across all datasets. More critically, this InceptionV3-based adapted IS failed to differentiate between the outputs of the coarse and fine-tuned generative models. As shown in Table 2, it yielded identical scores for both model versions across all datasets (e.g., TNBC: 1.00 for coarse vs. 1.00 for fine-tuned; DSB 2018: 1.05 vs. 1.05; MoNuSeg: 1.00 vs. 1.00; PanNuke: 1.07 vs. 1.07).

## Enhanced Evaluation using Domain-Specific Feature Extractors

When we replaced InceptionV3 with pathology-specific feature extractors (UNI, Virchow, Virchow 2), we observed a much wider range in FID scores. For example, UNI-based FID scores ranged from 962.81 to 2156.59, Virchow-based scores from 481.84 to 1369.57 and Virchow 2-based scores from 777.96 to 1971.40. These elevated scores suggest that domain-specific models may more effectively distinguish between real and synthetic data distribution in the histopathology domain. Importantly, pathology-specific FID metrics showed clearer decrease between coarse and fine-tuned which may indicate improvement patterns between coarse and finetuned models in most cases. For MoNuSeg, all three pathology-specific extractors showed improvement from coarse to finetuned model (UNI: 2156.59 to 1665.67, Virchow: 1369.57 to 1056.10, Virchow 2: 1971.40 to 1422.35), with percentage improvements of 22.8%, 22.9%, and 27.9% respectively.

Table 2 also shows that the pathology-specific models revealed more pronounced differences in IS between real and synthetic images compared with inception V3-based IS. While InceptionV3-based IS remained near 1.0, domain-specific IS values ranged from 1.17 to 1.65, providing greater numerical differentiation between coarse and fine-tuned image sets in this study. For example, UNI-based IS for DSB 2018 dataset reached 1.65, compared to just 1.05 with InceptionV3. The modified IS also showed increased discriminative properties to quality improvements between coarse and finetuned models in some cases, particularly with MoNuSeg (UNI: 1.44 to 1.50, Virchow: 1.22 to 1.25, Virchow 2: 1.27 to 1.32). However, in TNBC, we see the opposite trend, which we believe it is due to the quality of the original image dataset and generated images.

Table 2. Fréchet Inception Distance (FID) and Inception Score (IS) values obtained using different feature extractors. These metrics are evaluated for various synthetic datasets, with FID calculated between each synthetic dataset and the real dataset.

| | | **INCEPTION V3** | | **UNI** | | **VIRCHOW** | | **VIRCHOWV2** | |
|---|---|---|---|---|---|---|---|---|---|
| Dataset | | **FID↓** | **IS ↑** | **FID↓** | **IS ↑** | **FID↓** | **IS ↑** | **FID↓** | **IS ↑** |
| TNBC | Coarse | 70.36 | 1.00 | **1881.54** | **1.62** | 1334.13 | **1.28** | **1433.06** | **1.38** |
| | Finetune | **67.14** | 1.00 | 1935.34 | 1.56 | **1254.47** | 1.24 | 1608.15 | 1.34 |
| DSB 2018 | Coarse | 134.69 | 1.05 | 1276.77 | **1.65** | 629.57 | **1.34** | 885.50 | **1.33** |
| | Finetune | **123.64** | 1.05 | **1025.43** | 1.61 | **567.68** | 1.31 | **777.96** | 1.31 |
| MoNuSeg | Coarse | 41.72 | 1.00 | 2156.59 | 1.44 | 1369.57 | 1.22 | 1971.40 | 1.27 |
| | Finetune | **25.65** | 1.00 | **1665.67** | **1.50** | **1056.10** | **1.25** | **1422.35** | **1.32** |
| PanNuke | Coarse | 126.77 | 1.07 | 1163.27 | 1.59 | 646.79 | **1.21** | 1035.14 | 1.33 |
| | Finetune | **117.14** | 1.07 | **962.81** | 1.59 | **481.84** | 1.17 | **892.14** | **1.34** |

Table 3 reports the precision and recall for the different feature extractors estimated on the synthetic datasets. For the PanNuke dataset, InceptionV3 produced high precision values (coarse: 0.5538, finetuned: 0.6372) with a relatively small improvement between models (15.1%). This suggests that InceptionV3 perceives both model outputs as having high fidelity to real data, with minimal differentiation between quality levels. In contrast, all pathology specific feature extractors assigned lower precision values for the coarse compared to the fine-tuned for the same PanNuke dataset (UNI: 0.0097 to 0.0399, Virchow: 0.0834 to 0.1886, Virchow 2: 0.0097 to 0.0387). Furthermore, the pathology-specific extractors showed larger relative improvements between coarse and fine-tuned models (UNI: 311.3%, Virchow: 126.1%, VirchowV2: 298.9% increases), which indicates higher sensitivity to quality improvements.

Table 3. Precision and Recall values obtained using different feature extractors.

| | | **INCEPTION V3** | | **UNI** | | **VIRCHOW** | | **VIRCHOWV2** | |
|---|---|---|---|---|---|---|---|---|---|
| **Dataset** | | **Precision↑** | **Recall↑** | **Precision↑** | **Recall↑** | **Precision↑** | **Recall↑** | **Precision↑** | **Recall↑** |
| TNBC | Coarse | 0.6364 | 0.5455 | **0.4091** | **0.4545** | 0.2727 | **0.4091** | **0.2727** | **0.5909** |
| | Finetune | **0.7727** | **0.5909** | 0.2273 | 0.1364 | **0.4545** | 0.3182 | 0.1818 | 0.0909 |
| DSB 2018 | Coarse | 0.3056 | **0.6944** | 0.0833 | 0.5093 | 0.1852 | 0.6296 | 0.0926 | 0.5833 |
| | Finetune | **0.4907** | 0.6389 | **0.2222** | **0.7593** | **0.2130** | **0.7685** | **0.1667** | **0.6204** |
| MoNuSeg | Coarse | 0.1250 | 0.0625 | 0.6250 | 0.0000 | 0.9375 | 0.0625 | 0.5625 | 0.0 |
| | Finetune | **0.7500** | **0.6875** | **1.0000** | **0.3125** | **1.0000** | **0.2500** | **1.0000** | **0.375** |
| PanNuke | Coarse | 0.5538 | 0.1717 | 0.0097 | **0.0435** | 0.0834 | 0.0290 | 0.0097 | 0.0 |
| | Finetune | **0.6372** | **0.2503** | **0.0399** | 0.0375 | **0.1886** | **0.0846** | **0.0387** | **0.022** |

Dimensionality reduction visualizations are shown in Figure 3(a), revealing differences in how the feature extractors separate real and synthetic histopathology images. When using InceptionV3 features (pretrained on natural images), projections via PCA, t-SNE, and UMAP showed some overlap between real and synthetic distributions, potentially indicating only limited effect to domain-specific differences. In contrast, pathology-specific extractors (UNI, Virchow, Virchow2) produced more separated clusters, with the highest separation observed using the Virchow2 features. These results suggest that foundation models pretrained on digital pathology data may capture color bias and textural inconsistencies that InceptionV3 misses [37]. Nearest-neighbor analysis further highlighted these differences in Figure 3(b). InceptionV3 frequently matched images based on broad structural patterns, often pairing real and synthetic images with similar spatial layouts but differing staining characteristics. Pathology-specific models, however, consistently retrieved neighbors based on tissue morphology, cellular architecture, and clinically relevant pathology features. Pathologist review indicated that both model outputs displayed key histopathological structures, including carcinoma, stroma, and inflammatory/lymphocyte infiltrates, supporting their visual and clinical realism.

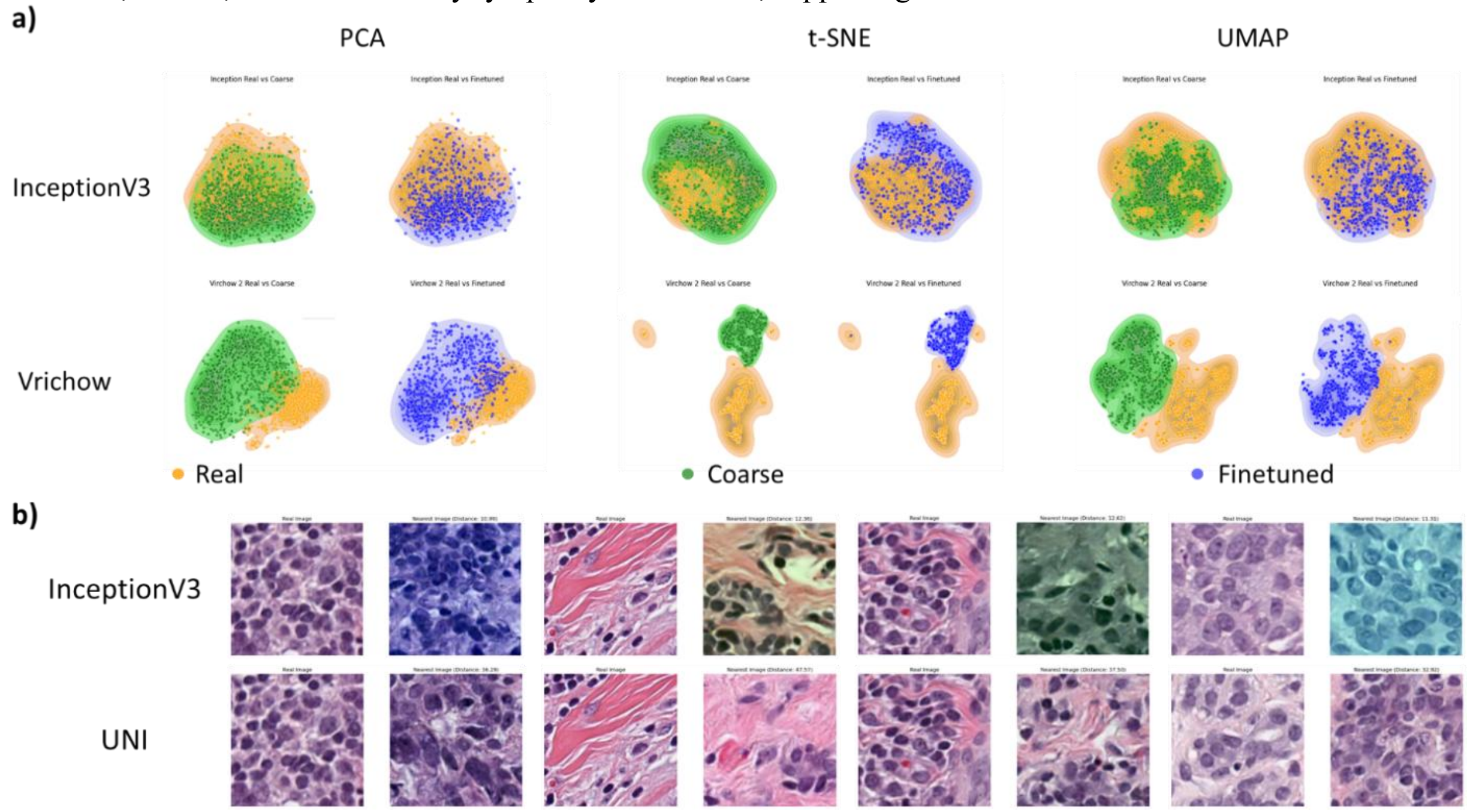

Figure 3. (a) Visualization of Feature Space Distribution. This figure visualizes the distribution of features extracted from real and synthetic histopathology images in a reduced-dimensional space. The visualizations are arranged in columns by dimension reduction technique: PCA (left), t-SNE (middle), and UMAP (right). Rows represent the feature extractor used: InceptionV3 (top row, pretrained on natural images) and Virchow (bottom row, pretrained on pathology images). Within each feature extractor's row, the left panel compares the feature distribution of real data from the coarse-trained model, while the right panel compares real data against synthetic data from the finetuned model. (b) The nearest neighbors calculated using different feature extractors. The top row displays results using InceptionV3 features, while the bottom row shows results using UNI features. In both rows, the image on the left of each pair is from the real dataset, and the image on the right is its corresponding nearest neighbor from the coarse model dataset.

## Downstream Task Analysis and Correlation Analysis

Table 4 shows segmentation performance test results (real images) for the HoVerNet model trained on different combinations of real and synthetic training data. Our experiments revealed consistent improvements in segmentation performance when augmenting real training data with fine-tuned (FT) synthetic images across all datasets (Table 4). For the MoNuSeg dataset, training with real and synthetic data produced a mean DICE of 0.7906 and AJI+ of 0.6347, representing improvements in DICE and AJI+ over training with real only images alone (DICE: 0.7790, AJI+: 0.6097). Similar positive trends were observed across other datasets but did not achieve statistically significant differences. These results suggest that fine-tuned synthetic data, which achieved higher pathology-specific metric scores, may provide improved augmentations for training downstream segmentation models.

Table 4. Segmentation testing result for the HoVerNet segmentation models trained on different numbers of real and synthetic data.

| | **Train (# images)** | **Tuning (# images)** | **Dice (mean, std)** | **AJI+ (mean, std)** |
|---|---|---|---|---|
| MoNuSeg | Real (16) | Real (14) | 0.7790 (0.0962) | 0.6097 (0.1039) |
| | Real (16) + FT (16) | Real (14) | **0.7906 (0.0982)** | **0.6347 (0.1063)** |
| PanNuke | Real (827) | Real (749) | 0.8358 (0.0852) | 0.6871 (0.1081) |
| | Real (827) + FT (827) | Real (749) | **0.8382 (0.0911)** | **0.6925 (0.1094)** |
| 2018 DSB | Real (108) | Real (16) | 0.7423 (0.02608) | 0.5208 (0.2010) |
| | Real (108) + FT (108) | Real (16) | **0.7883 (0.01842)** | **0.5635 (0.1670)** |
| TNBC | Real (22) | Real (18) | 0.7585 (0.0506) | 0.5297 (0.0588) |
| | Real (22) + FT (22) | Real (18) | **0.7667 (0.0414)** | **0.5463 (0.0560)** |

The Pearson correlation coefficients between the synthetic data quality metrics (FID and IS) and the resulting segmentation performance (AJI+ and Dice) are presented in Table 5. The analysis revealed that the modified Inception Score (IS), utilizing foundation model-based features, showed higher correlation with the downstream segmentation performance metrics, with a Pearson correlation of 0.6096 with AJI+ and 0.5639 with Dice. This finding may suggest that the modification to the Inception Score shows promise as an indicator of a synthetic dataset's potential utility for training on downstream tasks. The foundation model-based FID showed a negative correlation with Dice (-0.5597, p-value = 0.0242), which is consistent with lower FID, while the foundation-based IS provided a high positive correlation. Other metrics, including InceptionV3-based IS, precision, recall, density, and coverage, showed lower correlations with segmentation performance in this analysis. The observed higher correlation for the foundation model-based evaluation metrics supports the approach of using domain-specific feature extractors for evaluating synthetic histopathology images, offering a potential dataset metric to gauge whether a given generated synthetic dataset might be effective for downstream segmentation tasks, potentially reducing the need for extensive trial-and-error training of full downstream segmentation models for initial dataset assessment. However, it is important to note that these correlations are specific to the datasets, model architecture, and metrics used in this study and may not generalize to other contexts.

Table 5. Correlation estimates between the synthetic data evaluation metrics and the task-specific segmentation performance metrics.

| Rank | Performance Metric | Evaluation Metric | Pearson Correlation Coefficient | Pearson p-value |
|---|---|---|---|---|
| 1 | AJI+ | IS (Foundation Based) | 0.6096 | 0.0122 |
| 2 | Dice | IS (Foundation Based) | 0.5639 | 0.0229 |
| **3** | Dice | FID (Foundation Based) | -0.5597 | 0.0242 |
| **4** | AJI+ | FID (Foundation Based) | -0.4722 | 0.0648 |
| **5** | Dice | Density | 0.3597 | 0.1712 |
| **6** | AJI+ | Density | 0.2224 | 0.4077 |
| **7** | Dice | Recall | 0.2210 | 0.4107 |
| **8** | AJI+ | Precision | -0.1968 | 0.4650 |
| **9** | Dice | IS (InceptionV3 Based) | 0.1926 | 0.4748 |
| **10** | Dice | Coverage | 0.1393 | 0.6069 |
| **11** | AJI+ | IS (InceptionV3 Based) | 0.0708 | 0.7944 |
| **12** | AJI+ | Coverage | -0.0602 | 0.8247 |
| **13** | AJI+ | Recall | 0.0529 | 0.8458 |
| **14** | Dice | Precision | 0.0142 | 0.9584 |

## 4. CONCLUSIONS

In this study, we conducted an evaluation of synthetic histopathology images generated through conditional denoising diffusion models across multiple datasets (MoNuSeg, PanNuke, TNBC, and 2018 Data Science Bowl). This study focused on evaluating both traditional image quality metrics and domain-specific metrics and then assessing the correlation of these metrics with downstream nuclei segmentation performance metrics. The study involved evaluating synthetic images using both standard metrics (SSIM, FID, PSNR) and the pathology-specific modified FID and IS metrics. We generated synthetic images at varying fidelity levels (coarse and fine-tuned), trained nuclei segmentation models using these synthetic datasets, measured the segmentation performance via AJI+ and Dice scores, and calculating the statistical correlations between synthetic image quality metrics and segmentation performance metrics. The results suggest that the modified Inception Score using pathology-specific feature extractors shows highest observed correlation with the segmentation performance metrics (correlation of 0.6096 with AJI+ and 0.5639 with Dice), as well as for FID (-0.5597 AJI+ and -0.4722 Dice). The results suggest that domain-specific metrics may correlate more closely with downstream task performance than standard metrics, providing a potential tool for dataset assessment. for downstream computational pathology tasks. This may be especially important for rare morphological variants where collecting sufficient real-world training examples remains challenging. We performed our study on four different available nuclei segmentation datasets; however, these datasets have a limited number of annotated images since creating such datasets is resource-intensive and time-consuming. Our future work intends to further investigate the relationship between synthetic data attributes and downstream task performance, aiming to develop robust approaches for evaluating both fidelity and variety in synthetic pathology images.

## ACKNOWLEDGEMENT

The findings and conclusions in this article are those of the author(s) and do not necessarily represent the official position of the U.S. Food and Drug Administration (FDA), the Department of Health and Human Services (HHS), or the U.S. Government. Mention of trade names, commercial products, or organizations does not imply endorsement by the U.S. Government. This project was supported in part by the FDA's intramural Critical Path program and by an appointment to the ORISE Research Participation Program at the Center for Devices and Radiological Health, U.S. Food and Drug Administration, administered by the Oak Ridge Institute for Science and Education. We thank Dr. Christopher Trindade for reviewing the synthetic images and helpful discussions, and Dr. Alexej Gossmann, Dr. Arian Arab, and Dr. Ravi Samala for helpful discussions. We thank Tahsin Rahman and Alexander Webber for reviewing and editing papers.